\documentclass[conference]{IEEEtran}
\usepackage{times}

\usepackage[numbers]{natbib}
\usepackage{multicol}
\usepackage[bookmarks=true]{hyperref}

\usepackage{graphics} 
\usepackage{epsfig} 
\usepackage{mathptmx} 
\usepackage{times} 
\usepackage{amsmath} 
\usepackage{amssymb}  

\usepackage{bm}
\usepackage[short]{optidef}
\usepackage{algorithm}
\usepackage[noend]{algpseudocode}

\usepackage{tikzscale}
\usepackage{pgfplots}
\pgfplotsset{compat=1.9, legend style={font=\footnotesize}}
\usepgfplotslibrary{groupplots}

\usepackage{hyperref}
\hypersetup{
    colorlinks=true,
    linkcolor=blue,
    filecolor=magenta,      
    urlcolor=black,
}

\begin{document}
\renewcommand\UrlFont{\color{black}\ttfamily}

\title{ALGAMES: A Fast Solver for Constrained Dynamic Games}


\author{\authorblockN{Simon Le Cleac'h}
\authorblockA{Department of Mechanical Engineering\\
Stanford University\\
simonlc@stanford.edu}
\and
\authorblockN{Mac Schwager}
\authorblockA{Department of Aeronautics\\
\& Astronautics\\
Stanford University\\
schwager@stanford.edu}
\and
\authorblockN{Zachary Manchester}
\authorblockA{Department of Aeronautics \\
\& Astronautics\\
Stanford University\\
zacmanchester@stanford.edu}}


%

\maketitle

\begin{abstract}

Dynamic games are an effective paradigm for dealing with the control of multiple interacting actors. This paper introduces ALGAMES (Augmented Lagrangian GAME-theoretic Solver), a solver that handles trajectory optimization problems with multiple actors and general nonlinear state and input constraints. Its novelty resides in satisfying the first order optimality conditions with a quasi-Newton root-finding algorithm and rigorously enforcing constraints using an augmented Lagrangian formulation. We evaluate our solver in the context of autonomous driving on scenarios with a strong level of interactions between the vehicles. We assess the robustness of the solver using Monte Carlo simulations. It is able to reliably solve complex problems like ramp merging with three vehicles three times faster than a state-of-the-art DDP-based approach. A model predictive control (MPC) implementation of the algorithm demonstrates real-time performance on complex autonomous driving scenarios with an update frequency higher than 60 Hz.

\end{abstract}

\IEEEpeerreviewmaketitle

\section{Introduction}
    Controlling a robot in an environment where it interacts with other agents is a complex task. Traditional approaches in the literature adopt a predict-then-plan architecture. First, predictions of other agents' trajectories are computed, then they are fed into a planner that considers them as immutable obstacles. This approach is limiting because the effect of the robot's trajectory on the other agents is ignored. Moreover, it can lead to the ``frozen robot'' problem that arises when the planner finds that all paths to the goal are unsafe \cite{Trautman2010}. It is therefore crucial for a robot to \emph{simultaneously} predict the trajectories of other vehicles on the road while planning its own trajectory, in order to capture the reactive nature of all the agents in the scene.  ALGAMES provides such a joint trajectory predictor and planner by considering all agents as players in a Nash style dynamic game.  We envision ALGAMES as being run on-line by a robot in a receding horizon loop, at each iteration planning a trajectory for the robot by explicitly accounting for the reactive nature of all agents in its vicinity.

    Joint trajectory prediction and planning in scenarios with multiple interacting agents is well-described by a dynamic game.  Dealing with the game-theoretic aspect of multi-agent planning problems is a critical issue that has a broad range of applications. For instance, in autonomous driving, ramp merging, lane changing, intersection crossing, and overtaking maneuvers all comprise some degree of game-theoretic interactions \cite{Sadigh2016, Sadigh2016a, Fridovich-Keil2019a, Dreves2018, Fisac2019, Schmerling2018}. Other potential applications include mobile robots navigating in crowds, like package delivery robots, tour guides, or domestic robots; robots interacting with people in factories, such as mobile robots or fixed-base multi-link manipulators; and competitive settings like drone and car racing \cite{Spica2018, Liniger2019}.

    \begin{figure}[t]
    \centering
    \includegraphics[width=8.85cm]{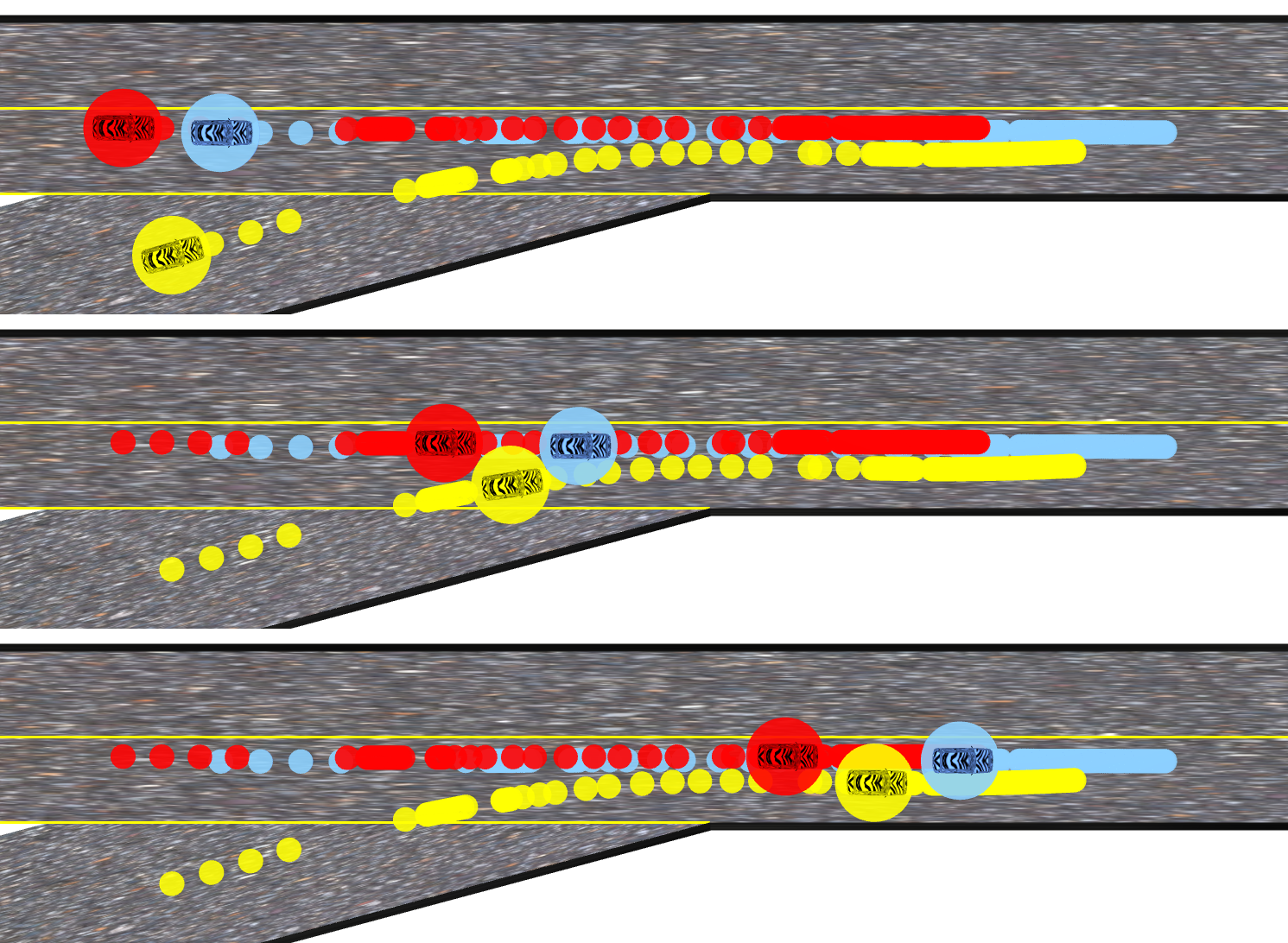}\hfill
    \caption{A ramp merging problem is solved using a receding horizon implementation of ALGAMES. Each time the receding-horizon plan is updated, we add a colored dot representing the position of each vehicle at the time of the update. In this simulation, ALGAMES has been called 151 times in 3 seconds to find receding horizon Nash equilibrium trajectories. ALGAMES generates seemingly natural trajectories that account for the individual objectives of all players while sharing the responsibility of avoiding collisions.}
    \label{fig:mpc_merging_dots}
    \end{figure}

    In this work, we seek solutions to constrained multi-player general-sum dynamic games. In dynamic games, the players' strategies are sequences of decisions. It is important to notice that, unlike traditional optimization problems, non-cooperative games have no ``optimal'' solution. Depending on the structure of the game, asymmetry between players, etc., different concepts of solutions are possible. In this work, we search for Nash equilibrium solutions. This type of equilibrium models symmetry between the players; all players are treated equally. At such equilibria, no player can reduce its cost by unilaterally changing its strategy. For extensive details about the game-theory concepts addressed in this paper, we refer readers to the work of Bressan \cite{Bressan2010} and Basar et al. \cite{Basar1999}.
    

    Our solver is aimed at finding a Nash equilibrium for multi-player dynamic games, and can handle general nonlinear state and input constraints. This is particularly important for robotic applications, where the agents often interact through their desire to avoid collisions with one another or with the environment. Such interactions are most naturally represented as (typically nonlinear) state constraints. This is a crucial feature that sets game-theoretic methods for robotics apart from game-theoretic methods in other domains, such as economics, behavioral sciences, and robust control. In these domains, the agent interactions are traditionally represented in the objective functions themselves, and these games typically have no state or input constraints. In mathematics literature, Nash equilibria with constraints are referred to as \emph{Generalized Nash Equilibria} \cite{Facchinei2007}.  Hence, in this paper we present an augmented Lagrangian solver for finding Generalized Nash Equilibria specifically tailored to robotics applications.  
    
    Our solver assumes that players are rational agents acting to minimize their costs. This rational behavior is formulated using the first-order necessary conditions for Nash equilibria, analogous to the Karush-Kuhn-Tucker (KKT) conditions in optimization. By relying on an augmented Lagrangian approach to handle constraints, the solver is able to solve multi-player games with several agents and a high level of interactions at real-time speeds. Finding a Nash equilibrium for 3 autonomous cars in a freeway merging scenario takes $90$ ms. Our primary contributions are:
    \begin{enumerate}
        \item A general solver for dynamic games aimed at identifying Generalized Nash Equilibrium strategies.
        \item A real time MPC implementation of the solver able to handle noise, disturbances, and collision constraints (Fig. \ref{fig:mpc_merging_dots}).
        \item A comparison with iLQGames \cite{Fridovich-Keil2019a} on speed. ALGAMES finds Nash equilibria 3 times faster than iLQGames for a fixed constraint satisfaction criterion.
    \end{enumerate}
\section{Related Work}

    \subsection{Equilibrium Selection}
\begin{figure}[t]
    \centering
    \includegraphics[width=8.85cm]{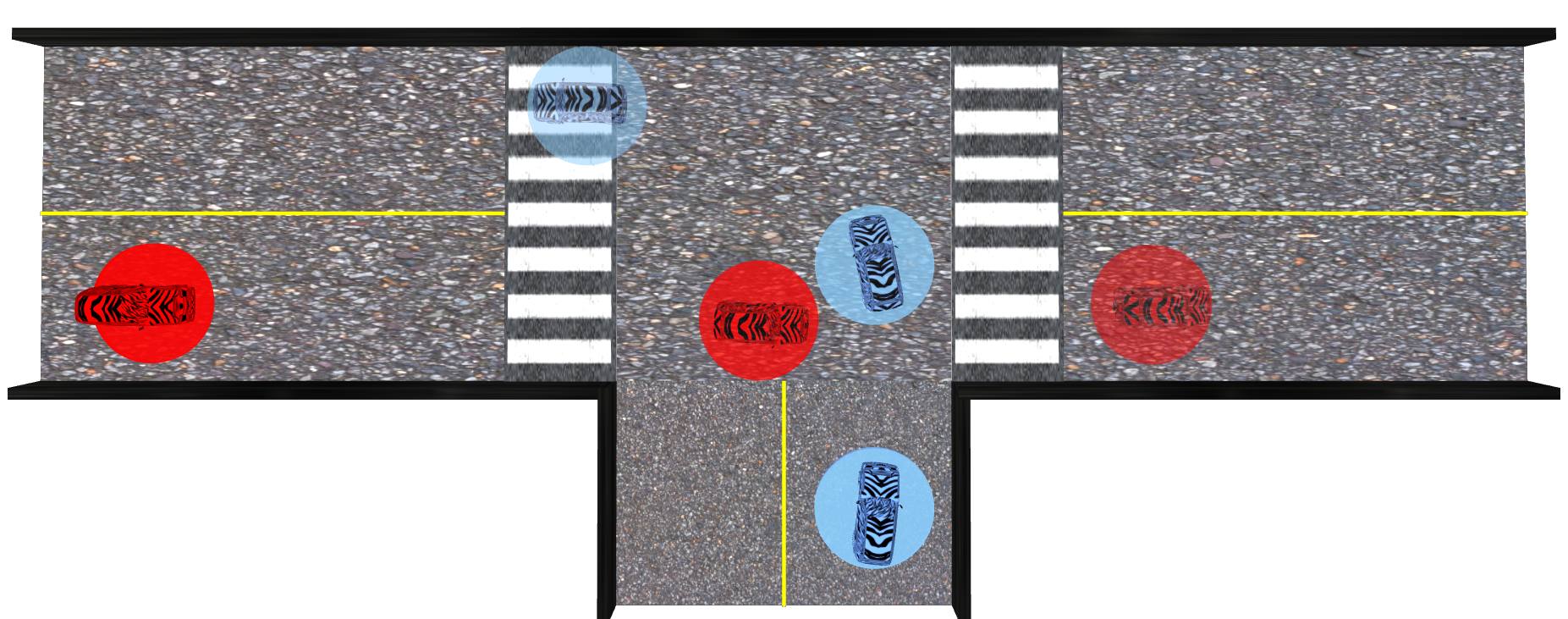}\hfill
    \caption{Superimposed sequence of images depicting the trajectories obtained by solving for open-loop Stackelberg equilibrium strategies. The slow blue vehicle is the leader and cuts in front of the the fast red vehicle, which is the follower. This example illustrates the fact that the leader always has the ``right of way'' over the follower in any situation.}
    \label{fig:stackelberg}
    \end{figure}

    Recent work focused on solving multi-player dynamic games can be categorized by the type of equilibrium they select. Several works \cite{Sadigh2016, Sadigh2016a, Liniger2019, Yoo2012} have opted to search for Stackelberg equilibria, which model an asymmetry of information between players. These approaches are usually formulated for games with two players, a leader and a follower. The leader chooses its strategy first, then the follower selects the best response to the leader's strategy. Alternatively, a Nash equilibrium does not introduce hierarchy between players; each player's strategy is the best response to the other players' strategies. As pointed out in \cite{Fisac2019}, searching for open-loop Stackelberg equilibrium strategies can fail on simple examples. In the context of autonomous driving, for instance, when players' cost functions only depend on their own state and control trajectories, the solution becomes trivial. The leader ignores mutual collision constraints and the follower has to adapt to this strategy. This behavior can be overly aggressive for the leader (or overly passive for the follower) and does not capture the game-theoretic nature of the problem, see Figure \ref{fig:stackelberg}. 
    
    Nash equilibria have been investigated in \cite{Fridovich-Keil2019a, Dreves2018, Spica2018, Britzelmeier2019, Di2018, Di2020, Di2020a}. We also take the approach of searching for Nash equilibria, as this type of equilibrium seems better suited to symmetric, multi-robot interaction scenarios. Indeed, we have observed more natural behavior emerging from Nash equilibria compared to Stackelberg when solving for open-loop strategies.

    \subsection{Game-Theoretic Trajectory Optimization}
    Most of the algorithms proposed in the robotics literature to solve for game-theoretic equilibria can be grouped into four types: First are algorithms aimed at finding Nash equilibria that rely on decomposition, such as Jacobi or Gauss-Siedel methods \cite{Spica2018, Britzelmeier2019, Wang2019a}. These algorithms are based on an iterative best response scheme in which players take turns at improving their strategies considering the other agents' strategies as immutable. This type of approach is easy to interpret and scales reasonably well with the number of players. However, convergence of these algorithms is not well understood \cite{Facchinei2007}, and special care is required to capture the game-theoretic nature of the problem \cite{Spica2018}.  Moreover, solving for a Nash equilibrium until convergence can require many iterations, each of which is a (possibly expensive) trajectory optimization problem. This can lead to prohibitively long solution times.
    
    Second, there are a variety of algorithms based on dynamic programming. In \cite{Fisac2019}, a Markovian Stackelberg strategy is computed via dynamic programming. This approach seems to capture the game-theoretic nature of autonomous driving. However, dynamic programming suffers from the curse of dimensionality, and therefore practical implementations rely on simplified dynamics models coupled with coarse discretization of the state and input spaces. To counterbalance these approximations, a lower-level planner informed by the state values under the Markovian Stackelberg strategy is run. This approach, which scales exponentially with the state dimension, has only been demonstrated in a two-player setting. Adding more players is likely to prevent real-time application of this algorithm. In contrast, our proposed approach scales polynomially with the number of players (see Section \ref{sec:complexity}).
    

    Third, algorithms akin to differential dynamic programming have been developed for robust control \cite{Morimoto2003} and later applied to game-theoretic problems \cite{Fridovich-Keil2019a, Di2018}. This approach scales polynomially with the number of players and is fast enough to run real-time in a model-predictive control (MPC) fashion \cite{Fridovich-Keil2019a}. However, this type of approach does not natively handle constraints. Collision-avoidance constraints are typically handled using large penalties that can result in numerical ill-conditioning which, in turn, can impact the robustness or the convergence rate of the solver. Moreover, it leads to a trade-off between trajectory efficiency and avoiding collisions with other players.
    
    Finally, algorithms that are analogous to direct methods in trajectory optimization have also been developed \cite{Di2020, Di2020a}. An algorithm based on a first-order splitting method was proposed by Di \cite{Di2020a} that is known to have a linear convergence rate. The experiments presented with this work show convergence of the algorithm after typically $10^3$ to $10^4$ iterations. A different approach based on Newton's method has been proposed \cite{Di2020}, but it is restricted to unconstrained dynamic games. Our solver belongs to this family of approaches. It also relies on a second-order Newton-type method, but it is able to handle general state and control input constraints. In addition, we demonstrate convergence on relatively complex problems in typically less than $10^2$ iterations.

    \subsection{Generalized Nash Equilibrium Problems}
    
    As mentioned above, we focus on finding Nash equilibria for multi-player games in which players are coupled through shared state constraints (such as collision-avoidance constraints). Therefore, these problems are instances of Generalized Nash Equilibrium Problems (GNEPs). The operations research field has a rich literature on GNEPs \cite{Pang2005, Facchinei2006, Facchinei2009, Facchinei2010, Fukushima2011}. Exact penalty methods have been proposed to solve GNEPs \cite{Facchinei2006, Facchinei2009}. Complex constraints such as those that couple players' strategies are handled using penalties, allowing solution of multi-player games jointly for all the players. However, these exact penalty methods require minimization of nonsmooth objective functions, which leads to slow convergence rates in practice.
    
    In the same vein, a penalty approach relying on an augmented Lagrangian formulation of the problem has been advanced by Pang et al. \cite{Pang2005}. This work, however, converts the augmented Lagrangian formulation to a set of KKT conditions, including complementarity constraints. The resulting constraint-satisfaction problem is solved with an off-the-shelf linear complementarity problem (LCP) solver that exploits the linearity of a specific problem. Our solver, in contrast, is not tailored for a specific example and can solve general GNEPs. It draws inspiration from the augmented Lagrangian formulation, which does not introduce nonsmooth terms in the objective function, enabling fast convergence. Moreover, this formulation avoids ill-conditioning, which improves the numerical robustness of our solver. 
    

\section{Problem Statement}
In the discretized trajectory optimization setting with $N$ time steps, we denote by $n$ the state size, $m$ the control input size, $x^{\nu}_k$ the state, and $u^{\nu}_k$  the control input of player $\nu$ at the time step $k$. In formulating the game, we do not distinguish between the robot carrying out the computation, and the other agents whose trajectories it is predicting.  All agents are modeled equivalently, as is typical in the case of Nash style games.

    Following the formalism of Facchinei \cite{Facchinei2007}, we consider the GNEP with $M$ players. Each player $\nu$ decides over its control input variables $U^{\nu} = [({u_1^{\nu}})^T \ldots ({u_{N-1}^{\nu}})^T]^T \in \mathbb{R}^{\bar{m}^\nu}$. This is player $\nu$'s strategy where $m^{\nu}$ denotes the dimension of the control inputs controlled by player $\nu$ and $\bar{m}^{\nu} = m^{\nu}(N-1)$ is the dimension of the whole trajectory of player $\nu$'s control inputs. By $U^{-\nu}$, we denote the vector of all the players' strategies except the one of player $\nu$. Additionally, we define the trajectory of state variables $X = [({x_2})^T \ldots ({x_{N}})^T]^T \in \mathbb{R}^{\bar{n}}$ where $\bar{n} = n(N-1)$, which results from applying all the control inputs decided by the players to a joint dynamical system,
    \begin{align}
        x_{k+1} = f(x_k, u^1_k, \ldots, u^M_k) = f(x_k, u_k),
    \end{align}
    with $k$ denoting the time step index. The kinodynamic constraints over the whole trajectory can be expressed with $\bar{n}$ equality constraints,
    \begin{align}
        D(X, U^1, \ldots, U^M) = D(X, U) = 0 \: \in \mathbb{R}^{\bar{n}}.
        \label{eq:dynamics}
    \end{align}
    The cost function of each player is noted $J^{\nu}(X, U^{\nu}): \mathbb{R}^{\bar{n}+\bar{m}^{\nu}} \rightarrow \mathbb{R}$. It depends on player $\nu$'s control inputs $U^{\nu}$ as well as on the state trajectory $X$, which is shared with all the other players. The goal of player $\nu$ is to select a strategy $U^{\nu}$ and a state trajectory $X$ that minimizes the cost function $J^{\nu}$. Naturally, the choice of state trajectory $X$ is constrained by the other players' strategies $U^{-\nu}$ and the dynamics of the system via Equation \ref{eq:dynamics}. In addition, the strategy $U^{\nu}$ must respect a set of constraints that depends on the state trajectory $X$ as well as on the other players strategies $U^{-\nu}$. We express this with a concatenated set of inequality constraints $C:\mathbb{R}^{\bar{n}+\bar{m}} \rightarrow \mathbb{R}^{n_c}$.
    Formally, 
    \begin{mini}[2]
    {X, U^{\nu}}{J^{\nu}(X, U^{\nu}),}{}{}
    \addConstraint{D(X, U) = 0}
    \addConstraint{C(X, U) \leq 0}
    \label{pb:gnep}.
    \end{mini}
    Problem (\ref{pb:gnep}), is a GNEP because of the constraints that couple the strategies of all the players. A solution of this GNEP (a generalized Nash equilibrium), is a vector $\hat{U}$ such that, for all $\nu = 1, \ldots, M$, $\hat{U}^{\nu}$ is a solution to (\ref{pb:gnep}) with the other players' strategies fixed to $\hat{U}^{-\nu}$. This means that at an equilibrium point $\hat{U}$, no player can decrease their cost by unilaterally changing their strategy $U^{\nu}$ to any other feasible point. 
    
    When solving for a generalized Nash equilibrium of the game, $U$, we identify open-loop Nash equilibrium trajectories, in the sense that the whole trajectory $U^{\nu}$ is the best response to the other players' strategies $U^{-\nu}$ given the initial state of the system $x_0$. Thus the control signal is a function of time, not of the current state of the system\footnote{One might also explore solving for feedback Nash equilibria, where the strategies are functions of the state of all agents.  This is an interesting direction for future work.} $x_k$. However, one can repeatedly resolve the open-loop game as new information is obtained over time to obtain a policy that is closed-loop in the model-predictive control sense, as demonstrated in Section \ref{sec:mpc}. This formulation is general enough to comprise multi-player general-sum dynamic games with nonlinear constraints on the states and control inputs. Practically, in the context of autonomous driving, the cost function $J^{\nu}$ encodes the objective of player $\nu$, while the concatenated set of constraints $C$ includes collision constraints coupled between players.

\section{Augmented Lagrangian Formulation}
    We propose an algorithm to solve the previously defined GNEP in the context of trajectory optimization. We express the condition that players are acting optimally to minimize their cost functions subject to constraints as an equality. To do so, we first derive the augmented Lagrangian associated with (\ref{pb:gnep}) solved by each player. Then, we use the fact that, at an optimal point, the gradient of the augmented Lagrangian is null \cite{Bertsekas2014}. Therefore, at a generalized Nash equilibrium point, the gradients of the augmented Lagrangians of all players must be null. Concatenating this set of $M$ equality constraints with the dynamics equality constraints, we obtain a set of equations that we solve using a quasi-Newton root-finding algorithm. 
    
    \setcounter{subsection}{0}
    \subsection{Individual Optimality}
    First, without loss of generality, we suppose that the vector $C$ is actually the concatenated set of inequality and equality constraints, i.e. $C = [C_i^T \, C_e^T]^T \in \mathbb{R}^{n_{ci}+n_{ce}}$, where $C_i \leq 0$ is the vector of inequality constraints and $C_e = 0$ is the vector of equality constraints. To embed the notion that each player is acting optimally, we formulate the augmented Lagrangian associated with (\ref{pb:gnep}) for player $\nu$. The dynamics constraints are handled with the Lagrange multiplier term $\mu^{\nu} \in \mathbb{R}^{\bar{n}}$,  while the other constraints are dealt with using both a multiplier and a quadratic penalty term specific to the augmented Lagrangian formulation. We denote by $\lambda \in \mathbb{R}^{n_c}$ the Lagrange multipliers associated with the vector of constraints $C$; $\rho \in \mathbb{R}^{n_c}$ is a penalty weight vector;
    \begin{align}
        L^{\nu}(X, U) = J^{\nu} + {\mu^{\nu}}^T D + {\lambda}^T C + \frac{1}{2} {C}^T I_{\rho}C .
        \label{eq:al}
    \end{align}
    $I_{\rho}$ is a diagonal matrix defined as, 
    \begin{align}
        I_{\rho,kk} &=
        \begin{cases}
            0            & \text{if} \:\:\: C_k(X,U) < 0 \: \land \: \lambda_k = 0, \: k \leq n_{ci}, \\
            \rho_k & \text{otherwise},
        \end{cases}
        \label{eq:penalty_logic}
    \end{align}
    where $k=1, \ldots, n_{ci}+n_{ce}$ indicates the $k^{\mathrm{th}}$ constraint. It is important to notice that the Lagrange multipliers $\mu^{\nu}$ associated with the dynamics constraints are specific to each player $\nu$, but the Lagrange multipliers and penalties $\lambda$ and $\rho$ are common to all players.
    Given the appropriate Lagrange multipliers $\mu^{\nu}$ and  $\lambda$, the gradient of the augmented Lagrangian with respect to the individual decision variables $\nabla_{X,U^{\nu}} \: L^{\nu} = G^{\nu}$ is null at an optimal point of (\ref{pb:gnep}). The fact that player $\nu$ is acting optimally to minimize $J^{\nu}$ under the constraints $D$ and $C$ can therefore be expressed as follows, 
    \begin{align}
        \nabla_{X, U^{\nu}} \: L^{\nu}(X, U, \mu^{\nu}) = G^{\nu}(X, U, \mu^{\nu}) = 0.
    \end{align}
    It is important to note that this equality constraint preserves coupling between players since the gradient $G^{\nu}$ depends on the other players' strategies $U^{-\nu}$. 
    
    \subsection{Root-Finding Problem}
    At a generalized Nash equilibrium, all players are acting optimally and the dynamics constraints are respected. Therefore, to find an equilibrium point, we have to solve the following root-finding problem,

    \begin{mini}[2]
    {X, U, \mu}{0, \quad \quad \quad \quad \quad \quad \quad \quad \quad \quad \quad \quad \quad \quad}{}{}
    \addConstraint{G^{\nu}(X, U, \mu^{\nu}) = 0, \:\:\: \forall \:\: \nu \in \{1, \ldots, M\}}
    \addConstraint{D(X,U) = 0}
    \label{pb:search},
    \end{mini}
    We use Newton's method to solve the root-finding problem. We denote by $G$ the concatenation of the augmented Lagrangian gradients of all players and the dynamics constraints, $G(X,U,\mu) = [({G^1})^T, \ldots, ({G^M})^T, D^T]^T $, where $\mu = [(\mu^1)^T, \ldots, (\mu^M)^T]^T \in \mathbb{R}^{\bar{n}M}$. We compute the first order derivative of $G$ with respect to the primal variables $X,U$ and the dual variables $\mu$ that we concatenate in a single vector $y = [(X)^T, (U)^T, (\mu)^{T}]$,
    \begin{align}
        H = \nabla_{X,U,\mu} G = \nabla_y G.
    \end{align}
    Newton's method allows us to identify a search direction $\delta y$ in the primal-dual space, 
    \vspace{-0.2cm}
    \begin{align}
        \delta y = - H^{-1}G.
        \label{eq:newton_step}
    \end{align}
    We couple this search direction with a backtracking line-search \cite{Nocedal2006} given in Algorithm \ref{al:linesearch} to ensure local convergence to a solution using Newton's Method \cite{Nocedal2006} detailed in Algorithm \ref{al:newton}.
    
    \begin{algorithm}
    \caption{Backtracking line-search}\label{al:linesearch}
    \begin{algorithmic}[1]
    \Procedure{LineSearch}{${y}, G, \delta{y}$}
    \State \textbf{Parameters}  
    \State $\alpha = 1$,
    \State $\beta \in (0, 1/2)$,
    \State $\tau \in (0, 1),$
    \State \textbf{Until} {$|| G({y}+\alpha\delta{y}) ||_1 < (1-\alpha\beta) || G({y}) ||_1$} \textbf{do} 
        \State \indent $\alpha \gets \tau \alpha$  
    \State \textbf{return} $\alpha$
    \EndProcedure
    \end{algorithmic}
    \end{algorithm}
    \vspace{-0.5cm}
    
    \begin{algorithm}
    \caption{Newton's method for root-finding problem}\label{al:newton}
    \begin{algorithmic}[1]
    \Procedure{Newton'sMethod}{${y}$}
    \State \textbf{Until} {Convergence} \textbf{do}
        \State \indent $G \gets [({\nabla_{X,U^{1}} \: L^{1}})^T, \ldots, ({\nabla_{X, U^{M}} \: L^{M}})^T, D^T]^T$ 
        \State \indent $H \gets \nabla_{y} G$ 
        \State \indent $\delta{y} \gets -H^{-1} G$ 
        \State \indent $\alpha \gets \Call{LineSearch}{{y}, G, \delta{y}}$
        \State \indent ${y} \gets {y} + \alpha \delta{y}$ 
    \State \textbf{return} ${y}$
    \EndProcedure
    \end{algorithmic}
    \end{algorithm}

    \begin{algorithm}
    \caption{ALGAMES solver}\label{al:algames}
    \begin{algorithmic}[1]
    \Procedure{ALGAMES}{${y}_0, \rho_0$}
    \State \textbf{Initialization}  
    \State $\rho \gets \rho^{(0)}, $
    \State $\lambda \gets 0, $   
    \State $\mu^{\nu} \gets 0, $   \hspace*{\fill} $\forall \nu$ 
    \State ${X, U} \gets X^{(0)}, U^{(0)}$ 
    \State \textbf{Until} {Convergence} \textbf{do}
        \State \indent ${y} \gets \Call{Newton'sMethod}{{y}} $ 
        \State \indent $\lambda \gets \Call{DualAscent}{{y}, \lambda, \rho},$
        \State \indent $\rho \gets \Call{IncreasingSchedule}{\rho},$ 
    \State \textbf{return} ${y}$
    \EndProcedure
    \end{algorithmic}
    \end{algorithm}

    \subsection{Augmented Lagrangian Updates}
    To obtain convergence of the Lagrange multipliers $\lambda$, we update them with a dual-ascent step. This update can be seen as shifting the value of the penalty terms into the Lagrange multiplier terms, 
    \begin{align}
        \lambda_k \leftarrow
        \begin{cases}
            \max(0, \lambda_k + \rho_k C_k(X, U)) 
                & k \leq n_{ci}, \\
            \lambda_k + \rho_k C_k(X,U)   
                & n_{ci} < k \leq n_{ci}+n_{ce}.
        \end{cases}
        \label{eq:dual_ascent}
    \end{align}
    We also update the penalty weights according to an increasing schedule, with $\gamma > 1$:
    \begin{align}
        \rho_k \leftarrow \gamma \rho_k, \:\:\: \forall k \in \{1, \ldots, n_c\}.
        \label{eq:pen_update}
    \end{align}

    \subsection{ALGAMES}
    By combining Newton's method for finding the point where the dynamics is respected and the gradients of the augmented Lagrangians are null with the Lagrange multiplier and penalty updates, we obtain our solver ALGAMES (Augmented Lagrangian GAME-theoretic Solver) presented in Algorithm \ref{al:algames}. The algorithm, which iteratively solves the GNEP, requires as inputs an initial guess for the primal-dual variables $y^{(0)}$ and initial penalty weights $\rho^{(0)}$. The algorithm outputs the open-loop strategies of all players $X,U$ and the Lagrange multipliers associated with the dynamics constraints $\mu$. 
    
    \subsection{Algorithm Complexity} \label{sec:complexity}
    Following a quasi-Newton approximation of the matrix $H$ \cite{Nocedal2006}, we neglect some of the second-order derivative terms associated with the constraints. Therefore, the most expensive part of the algorithm is the Newton step defined by Equation \ref{eq:newton_step}. By exploiting the sparsity pattern of the matrix $H$, we can solve Equation \ref{eq:newton_step} in $O(N(n+m)^3)$. Indeed, the sparsity structure allows us to perform a back-substitution scheme akin to solving a Riccati equation, which has known complexity of $O(N(n+m)^3)$. The complexity is cubic in the number of states $n$ and the number of control inputs $m$, which are typically linear in the number of players $M$. Therefore, the overall complexity of the algorithm is $O(N M^3)$. 

    \subsection{Algorithm Discussion}
    Here we discuss the inherent difficulty in solving for Nash equilibria in large problems, and explain some of the limitations of our approach. First of all, finding a Nash equilibrium is a non-convex problem in general. Indeed, it is known that even for single-shot discrete games, solving for exact Nash equilibria is computationally intractable for a large number of players    \cite{DaskalakisEtAlSIAMJournalonComputing08ComplexityOfNash}.  It is therefore not surprising that, in our more difficult setting of a dynamic game in continuous space, no guarantees can be provided about finding an exact Nash equilibrium.  Furthermore, in complex interaction spaces, constraints can be highly nonlinear and nonconvex. This is the case in the autonomous driving context with collision avoidance constraints. In this setting, one cannot expect to find an algorithm working in polynomial time with guaranteed convergence to a Nash equilibrium respecting constraints. On the other hand, \emph{local} convergence of Newton's method to open-loop Nash equilibria (that is, starting sufficiently close to the equilibrium, the algorithm will converge to it) has been established in the unconstrained case \cite{Di2020}. Our approach solves a sequence of unconstrained problems via the augmented Lagrangian formulation. Each of these problems, therefore, has guaranteed \emph{local} convergence. However, the overall method has no guarantee of global convergence to a generalized Nash equilibrium, and this is expected given the known computational difficulty of the problem.  
    
    Second, our algorithm requires an initial guess for the state and control input trajectories $X$, $U$ and the dynamics multipliers $\mu$. Empirically, we observe that choosing $\mu = 0$ and simply rolling out the dynamics starting from the initial state $x_0$ without any control was a sufficiently good initial guess to get convergence to a local optimum that respects both the constraints and the first-order optimality conditions. For a detailed empirical study of the convergence of ALGAMES and its failure cases, we refer to Sections \ref{sec:monte_carlo} and \ref{sec:failure}. 

    Finally, even for simple linear quadratic games, the Nash equilibrium solution is not necessarily unique. In general, an entire subspace of equilibria exists. In this case, the matrix $H$ in Equation \ref{eq:newton_step} will be singular. In practice, we regularize this matrix so that large steps $\delta y$ are penalized, resulting in an invertible matrix $H$ and convergence to a Nash equilibrium that minimizes the norm of $y$.
    

\section{Simulations: Design and Setup}
    We choose to apply our algorithm in the autonomous driving context. Indeed, many maneuvers like lane changing, ramp merging, overtaking, and intersection crossing involve a high level of interaction between vehicles. We assume a single car is computing the trajectories for all cars in its neighborhood, so as to find its own trajectory to act safely among the group. We assume that this car has access to a relatively good estimate of the surrounding cars' objective functions. Such an estimate could, in principle, be obtained by applying inverse optimal control on observed trajectories of the surrounding cars. 
    
    In a real application, the car could be surrounded by other cars that might not necessarily follow a Nash equilibrium strategy. In this case, we demonstrate empirically that by repeating the computation as frequently as possible in an MPC fashion, we obtain safe and adaptive autonomous behaviors.

    

    \setcounter{subsection}{0}
    \subsection{Autonomous Driving Problem}
    \subsubsection{Constraints}
    Each vehicle in the scene is an agent of the game. Our objective is to find a generalized Nash equilibrium trajectory for all of the vehicles. These trajectories have to be dynamically feasible. The dynamics constraints at time step $k$ are expressed as follows, 
    \begin{align}
        x_{k+1} = f(x_k, u^{1}_k, \ldots, u^{M}_k).
    \end{align}
    We consider a nonlinear unicycle model for the dynamics of each vehicle. A vehicle state, $x^{\nu}_k$, is composed of a 2D position, a heading angle and a scalar velocity. The control input $u^{\nu}_k$ is composed of an angular velocity and a scalar acceleration. 
    
    In addition, it is critical that the trajectories respect collision-avoidance constraints. We model the collision zone of the vehicles as circles of radius $r$. The collision constraints between vehicles are then simply expressed in terms of the position $\tilde{x}^{\nu}_{k}$ of each vehicle, 
    \begin{align}
        r^2 - || \tilde{x}^{\nu}_{k} - \tilde{x}^{\nu'}_{k} ||_2^2 \leq 0, \quad \forall \: \: \nu, \nu' \in \{1, \ldots, M\},  \nu \neq \nu'.
    \end{align} 
    We also model boundaries of the road to force the vehicles to remain on the roadway. This means that the distance between the vehicle and the closest point, $q$, on each boundary, $b$, has to remain larger than the collision circle radius, $r$,
    \begin{align}
        r^2 - || \tilde{x}^{\nu}_{k} - q_b||_2^2 \leq 0, \quad \forall \:\: b, \: \forall \: \: \nu \in \{1, \ldots, M\}.
    \end{align}

    In summary, based on reasonable simplifying assumptions, we have expressed the driving problem in terms of non-convex and non-linear coupled constraints.  
    
    \subsubsection{Cost Function}
    We use a quadratic cost function penalizing the use of control inputs and the distance between the current state and the desired final state $x_f$ of the trajectory,
    \begin{align}
        J^{\nu}(X, U^{\nu}) = &\sum_{k=1}^{N-1} \frac{1}{2}(x_k - x_f)^T Q (x_k - x_f) + \frac{1}{2}{u^{\nu}_k}^T R u^{\nu}_k + \\
        &\frac{1}{2}(x_N - x_f)^T Q_f (x_N - x_f).
    \end{align}
    This cost function only depends on the decision variables $p^{\nu}$ of vehicle $\nu$. Players' behaviors are coupled only through collision constraints. We could also add terms depending on other vehicles' strategies, such as a congestion penalty. 

\section{Comparison to iLQGames}
    \subsection{Motivation}
    In order to evaluate the merits of ALGAMES, we compare it to iLQGames \cite{Fridovich-Keil2019a} which is a DDP-based algorithm for solving general dynamic games. Both algorithms solve the problem by iteratively solving linear-quadratic approximations that have an analytical solution \cite{Basar1999}. For iLQGames, the augmented objective function $\hat{J}^{\nu}$ differs from the objective function, $J^{\nu}$, by a quadratic term penalizing constraint violations,
    \begin{align}
        \hat{J}^{\nu}(X,U) = J^{\nu}(X,U) + \frac{1}{2} C(X,U)^T I_{\rho} C(X,U).
    \end{align}
    Where $I_\rho$ is defined by, 
    \begin{align}
        I_{\rho,kk} &=
        \begin{cases}
            0            & \text{if} \:\:\: C_k(X,U) < 0 , \: k \leq n_{ci}, \\
            \rho_k & \text{otherwise}.
        \end{cases}
    \end{align}
    Here $\rho$ is an optimization hyperparameter that we can tune to satisfy constraints. For ALGAMES, the augmented objective function, $L^{\nu}$, is actually an augmented Lagrangian, see Equation \ref{eq:al}. The hyperparameters for ALGAMES are the initial value of $\rho^{(0)}$ and its increase rate $\gamma$ defined in Equation \ref{eq:pen_update}.

    \begin{figure}[t]
    \centering
    \includegraphics[width=8.85cm]{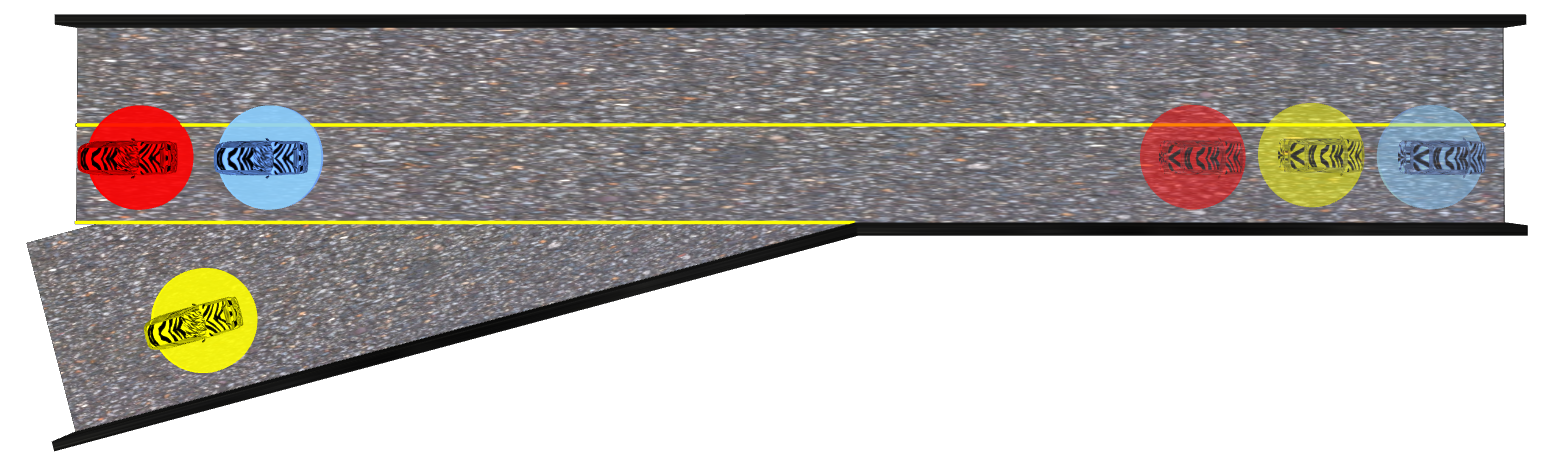}\hfill
    \caption{On the left, the three cars at their nominal initial state. On the right, the three cars are faded and standing at the nominal desired final state. The yellow car has successfully merged in between the two other cars. The roadway boundaries are depicted in black.}
    \label{fig:ramp_merging}
    \end{figure}

    \begin{figure}[t]
        \begin{center}
        \begin{tabular}{|c|c|c|c|}
        \hline
        Scenario & \# Players & ALGAMES & iLQGames\\
        \hline
          & 2 & $\mathbf{38 \pm 10 ms}$ & $104 \pm 23 ms$\\
        Ramp merging & 3 & $\mathbf{89 \pm 14 ms}$ & $197 \pm 15 ms$\\
          & 4 & $860 \pm 251 ms$ & $\mathbf{705 \pm 209 ms}$\\
        \hline
          & 2 & $\mathbf{50 \pm 11 ms}$ & $752 \pm 168 ms$\\
        Intersection & 3 & $\mathbf{116 \pm 22 ms}$ & $362 \pm 93 ms$\\
          & 4 & $\mathbf{509 \pm 33 ms}$ & $1905 \pm 498 ms$\\
        \hline
        \end{tabular}
        \caption{For each scenario and each number of players, we run each solver 100 times to estimate the mean solve time and its standard deviation.}
        \label{fig:benchmark}
        \end{center}
    \end{figure}

    \subsection{Timing Experiments}
\begin{figure}[t]
    \centering
    \includegraphics[width=8.85cm]{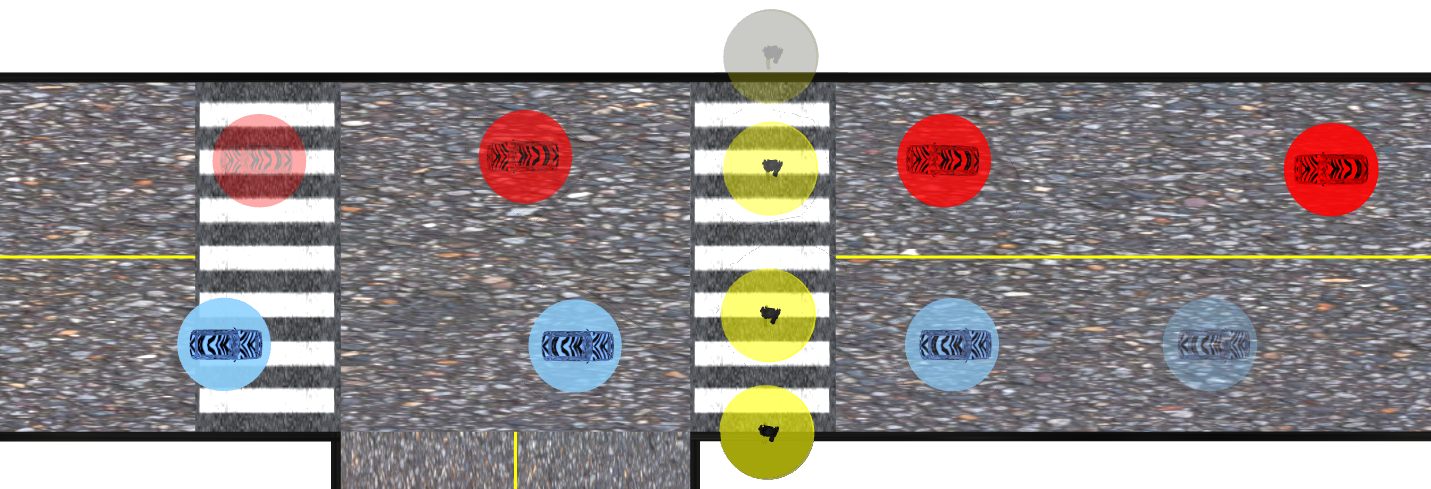}\hfill
    \caption{The blue car starts on the left and finishes on the right. The red does the opposite. The pedestrian with the yellow collision avoidance cylinder crosses the road from the bottom to the top of the image.}
    \label{fig:intersection}
    \end{figure}

    We evaluate the performance of both algorithms in two scenarios, see Figures \ref{fig:ramp_merging} and \ref{fig:intersection}, with the number of players varying from two to four. To compare the speed of both algorithms, we set the termination criterion as a threshold on constraint violations $C \leq 10^{-3}$. The timing results averaged over 100 samples are presented in Table \ref{fig:benchmark}. First, we notice that both algorithms achieve real-time or near-real-time performance on complex autonomous driving scenarios (the horizon of the solvers is fixed to $5s$).
    
    We observe that the speed performance of ALGAMES and iLQGames are comparable in the ramp merging scenario. For this scenario, we tuned the value of the penalty for iLQGames to $\rho = 10^2$. Notice that for all scenarios the dimensions of the problem are scaled so that the velocities and displacements are all the same order of magnitude. For the intersection scenario, we observe that the two-player and four-player cases both have much higher solve times for iLQGames compared to the 3-player case. Indeed, in those two cases, we had to increase the penalty to $\rho = 10^3$, otherwise the iLQGames would plateau and never reach the constraint satisfaction criterion. This, in turn, slowed the algorithm down by decreasing the constraint violation convergence rate. 
    
    \subsection{Discussion}
    The main takeaway from these experiments is that, for a given scenario, it is generally possible to find a suitable value for $\rho$ that will ensure the convergence of iLQGames to constraint satisfaction. With higher values for $\rho$, we can reach better constraint satisfaction at the expense of slower convergence rate. In the context of a receding horizon implementation (MPC), finding a good choice of $\rho$ that would suit the whole sequence of scenarios encountered by a vehicle could be difficult. In contrast, the same hyperparameters $\rho^{(0)}=1$ and $\gamma =10$ were used in ALGAMES for all the experiments across this paper. This supports the idea that, thanks to its adaptive penalty scheme, ALGAMES requires little tuning. 
    
    While performing the timing experiments, we also noticed several instances of oscillatory behavior for iLQGames. The solution would oscillate, preventing it from converging. This happened even after an adaptive regularization scheme was implemented to regularize iLQGames' Riccati backward passes. Oscillatory behavior was not seen with ALGAMES. We hypothesize that this is due to the dual ascent update coupled with the penalty logic detailed in Equations \ref{eq:dual_ascent} and \ref{eq:penalty_logic}, which add hysteresis to the solver.

    \begin{figure}
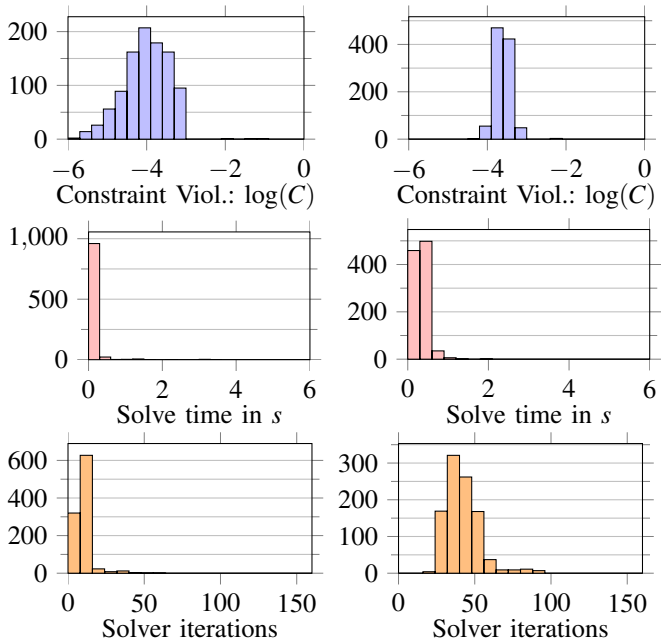

        \includegraphics[width=.49\linewidth, height=.32\linewidth]{tikz_hist/const_dir.tikz}\hfill
        \includegraphics[width=.49\linewidth, height=.32\linewidth]{tikz_hist/const_ilq.tikz}
        \includegraphics[width=.49\linewidth, height=.32\linewidth]{tikz_hist/time_dir.tikz}\hfill
        \includegraphics[width=.49\linewidth, height=.32\linewidth]{tikz_hist/time_ilq.tikz}
        \includegraphics[width=.49\linewidth, height=.32\linewidth]{tikz_hist/iter_dir.tikz}\hfill
        \includegraphics[width=.49\linewidth, height=.32\linewidth]{tikz_hist/iter_ilq.tikz}
        \caption{Monte Carlo analysis with 1000 randomly sampled initial states of ALGAMES on the left and iLQGames on the right. The top and middle plots indicate maximum constraint violation of the solution at the end of the solve and the solve time respectively. The bottom left and right plots displays the number of Newton steps and the number of Riccati backward passes executed during the solve of ALGAMES and iLQGames respectively.}
        
        \label{fig:monte_carlo}
    \end{figure}
    
    

    \begin{figure}[t]
        \begin{center}
        \begin{tabular}{|c c c c|} 
            \hline
            Scenario     & Freq. in Hz & $\mathbb{E}[\delta t]$ in ms & $\mathbb{\sigma}[\delta t]$ in ms \\ [0.5ex] 
            \hline
            Ramp Merging & 69 & 14 & 72 \\ 
            Intersection & 66 & 15 & 66 \\
            \hline
        \end{tabular}
        \caption{Running the MPC implementation of ALGAMES 100 times on both scenarios, we obtain the mean update frequency of the MPC as well as the mean and standard deviation of $\delta t$, the time required to update the MPC plan.}
        \label{fig:mpc_table}
        \end{center}
    \end{figure}

    \subsection{Monte Carlo Analysis} \label{sec:monte_carlo}
    To evaluate the robustness of ALGAMES, we performed a Monte Carlo analysis of its performance on a ramp merging problem. First, we set up a roadway with hard boundaries as pictured in Fig. \ref{fig:ramp_merging}. We position two vehicles on the roadway and one on the ramp in a collision-free initial configuration. We choose a desired final state where the incoming vehicle has merged into the traffic. Our objective is to generate generalized Nash equilibrium trajectories for the three vehicles. These trajectories are collision-free and cannot be improved unilaterally by any player. To introduce randomness in the solving process, we apply a random perturbation to the initial state of the problem. Specifically, we perturb $x_0$ by adding a uniformly sampled noise. This would typically correspond to displacing the initial position of the vehicles by $\pm 1 m$, changing their initial velocity by $\pm 3\%$ and their heading by $\pm 2.5^\circ$. 
    
    We observe in Figure \ref{fig:monte_carlo}, that ALGAMES consistently finds a satisfactory solution to the problem using the same hyperparameters $\rho^{(0)} =1$ and $\gamma = 10$. Out of the 1000 samples $99.5\%$ converged to constraint satisfaction $C \leq 10^{-3}$ while respecting the optimality criterion $||G||_1 < 10^{-2}$. By definition, $||G||_1$ is a merit function for satisfying optimality and dynamics constraints.  We also observe that the solver converges to a solution in less than $0.2 s$ for $96\%$ of the samples. The solver requires less than 16 Newton steps to converge for $94\%$ of the samples. These empirical data tend to support the fact that ALGAMES is able to solve the class of ramp merging problem quickly and reliably. 
    
    For comparison, we present in Figure \ref{fig:monte_carlo} the results obtained with iLQGames. We apply the same constraint satisfaction criterion $C \leq 10^{-3}$. We fixed the value of the penalty hyperparameter $\rho$ for all the samples as it would not be a fair comparison to tune it for each sample. Only 3 samples did not converge with iLQGames, this is a performance comparable to ALGAMES for which 5 samples failed to converge. However, we observe that iLQGames is 3 times slower than ALGAMES with an average solve time of $350$ ms compared to $110$ ms and require on average 4 times more iterations (9 against 41).

    
    \subsection{Solver Failure Cases}  \label{sec:failure}
    The Monte Carlo analysis allows us to identify the typical failure cases of our solver. We empirically identify the cases where the solver does not satisfy the constraints or the optimality criterion for the ramp merging problem. Typically in such cases, the initial guess, which consists of rolling out the dynamics with no control, is far from a reasonable solution. Since the constraints are ignored during this initial rollout, the car at the back can overtake the car at the front by driving through it. This creates an initial guess where constraints are strongly violated. Moreover, we hypothesize that the tight roadway boundary constraints tend to strongly penalize solutions that would 'disentangle' the car trajectories because they would require large boundary violation at first. Therefore, the solver gets stuck in this local optimum where cars overlap each other. Sampling several initial guesses with random initial control inputs and solving in parallel could reduce the occurrence of these failure cases. Also being able to detect, reject and re-sample initial guesses when the initial car trajectories are strongly entangled could also improve the robustness of the solver.

\section{MPC Implementation of ALGAMES} \label{sec:mpc}
    In this section, we propose a model-predictive control (MPC) implementation of the algorithm that demonstrates real-time performance. The benefits of the MPC are twofold: it provides a feedback policy instead of an open-loop strategy, and it can improve interactions with actors for which we do not have a good estimate of the objective function. 
    
    \subsection{MPC Feedback Policy}
    First, the strategies identified by ALGAMES are open-loop Nash equilibrium strategies. They are sequences of control inputs. On the contrary, DDP-based approaches like iLQGames, solve for feedback Nash equilibrium strategies which provide a sequence of control gains.
    In the MPC setting, we can obtain a feedback policy with ALGAMES by updating the strategy as fast as possible and only executing the beginning of the strategy. This assumes a fast update rate of the solution. To support the feasibility of the approach, we implemented an MPC on the ramp merging scenario described in Figure \ref{fig:ramp_merging}. There are 3 players constantly maintaining a 40 time step strategy with 3 seconds of horizon. We simulate 3 seconds of operation of the MPC by constantly updating the strategies and propagating noisy unicycle dynamics for each vehicle. We compile the results from 100 MPC trajectories in Table \ref{fig:mpc_table}. We obtain a $69$ Hz update frequency for the planner on average. We observe similar performance on the intersection problem defined in Figure \ref{fig:intersection}, with an update frequency of $66$ Hz. 
    

    \begin{figure}[t]
    \centering
    \includegraphics[width=8.85cm]{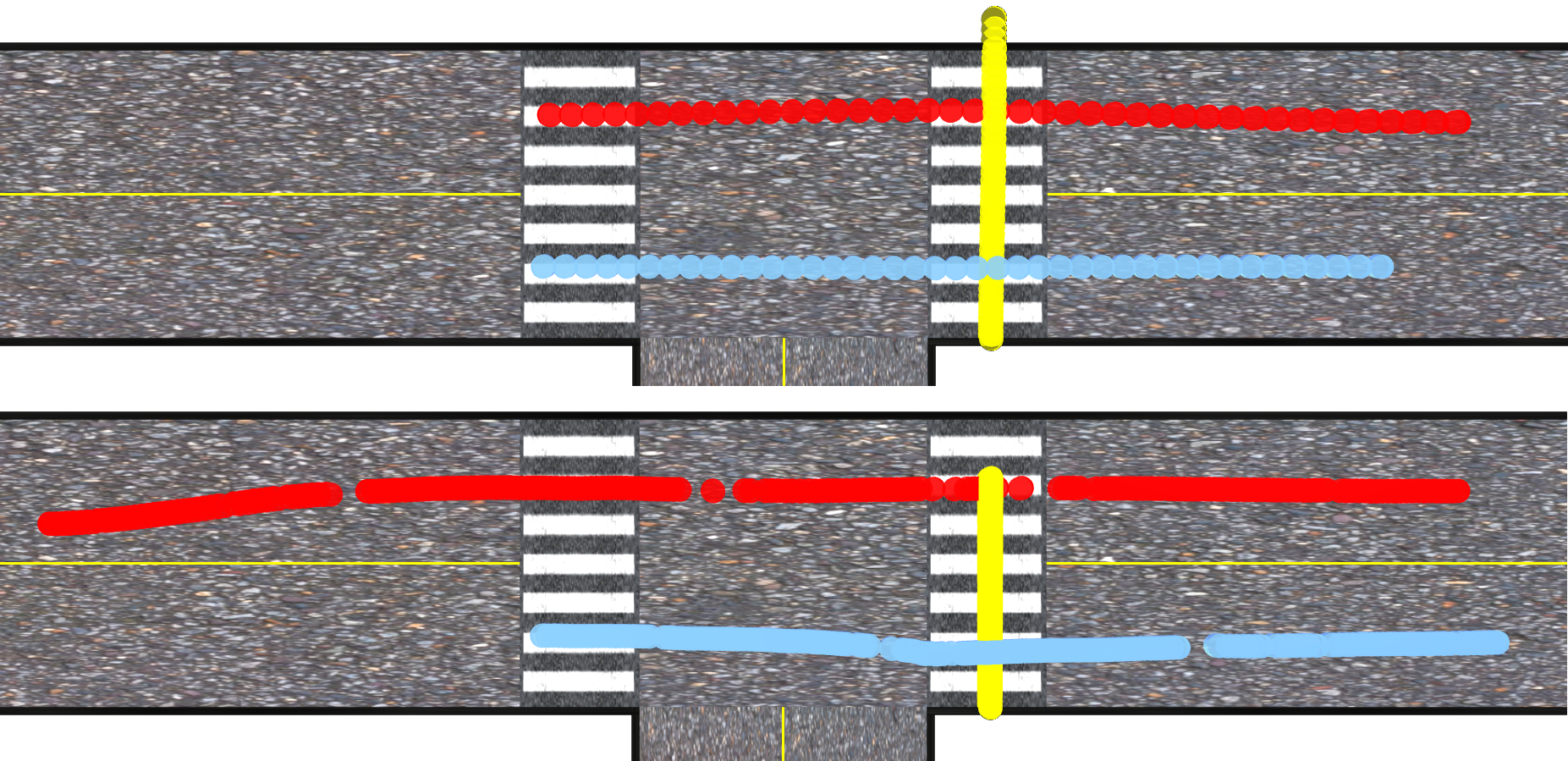}\hfill
    \includegraphics[width=8.85cm, height=3.3cm]{tikz_figures/velocity_mpc_pedestrian.tikz}\hfill
    \caption{The top plot represents the initial strategy that the blue car computes for all the players. There is one color dot for each players at each step of the planned strategy. Notice that the blue car predicts that following a straight path at a relatively constant speed will avoid collision with the yellow pedestrian. The middle chart shows the actual path of the 3 players by plotting their positions for each MPC strategy update. Notice that the blue car has to nudge to avoid the yellow pedestrian. The bottom plot shows that the blue car also dramatically slows down compared to the initial plan, to avoid colliding with the yellow pedestrian.}
    \label{fig:mpc_pedestrian}
    \end{figure}

    \subsection{Adaptive Behavior}
    The second benefit of MPC is that it mitigates a major assumption of this work. We assumed that the car we control has access to the exact objective functions of the surrounding cars. A more realistic setting would be that the car has access to a noisy estimate of the surrounding cars' objective functions. This estimate could be provided by an inverse optimal control algorithm, for instance. By re-planning at a high enough frequency, an MPC implementation of ALGAMES could safely control an agent who has an inaccurate estimate of surrounding agents' objective functions. We further assumed that the other players solve for Nash equilibrium strategies, which is not necessarily the case in the presence of selfish players.
    
    To support this claim, we modified the intersection scenario described in Figure \ref{fig:intersection} so that the blue car has a poor estimate of the pedestrian desired speed. Specifically, in its nominal strategy, the blue car crosses the intersection without slowing down crossing the crosswalk right after the yellow pedestrian. For this nominal strategy the blue car has solved for the GNE assuming that the pedestrian would have a desired speed $v_d$. However, in reality, the pedestrian is crossing the road at speed $v_0$ that is significantly lower than $v_d$. In addition, the pedestrian is not solving for any GNE and is just crossing the road in a straight line at a constant speed $v_0$. Therefore, the pedestrian objective function assumed by the blue car does not capture the real behavior of the pedestrian. However, when applying the MPC implementation of ALGAMES, we observe that the blue car gradually adapts its strategy to accommodate for the pedestrian, see Figure \ref{fig:mpc_pedestrian}. Indeed, as the blue car gets closer to the pedestrian the car significantly slows down compared to the nominal strategy. Also, we observe that the car shifts to the right of the roadway to avoid the pedestrian. This nudging maneuver was not present in the nominal plan because the pedestrian was expected to have crossed the lane already. It is important to note that this adaptive behavior is observed even though the blue car kept a constant and wrong estimate of the pedestrian's objective function. Being able to refine the estimates of other players objectives online could further improve the adaptive property of the algorithm.

\section{Conclusions}
    We have introduced a new algorithm for finding constrained Nash equilibrium trajectories in multi-player dynamic games. We demonstrated the performance and robustness of the solver through a Monte Carlo analysis on complex autonomous driving scenarios including nonlinear and non-convex constraints. We have shown real-time performance for up to 4 players and implemented ALGAMES in a receding-horizon framework to give a feedback policy. We empirically demonstrated the ability to safely interact with players that violate the Nash equilibrium assumptions when the strategies are updated fast enough online. The results we obtained from ALGAMES are promising, as they seem to let the vehicles share the responsibility for avoiding collisions, leading to natural-looking trajectories where players are able to negotiate complex, interactive traffic scenarios that are challenging for traditional, non-game-theoretic trajectory planners. For this reason, we believe that this solver could be a very efficient tool to generate trajectories in situations where the level of interaction between players is strong. Our implementation of ALGAMES is available at \url{https://github.com/RoboticExplorationLab/ALGAMES.jl}.

\bibliographystyle{ieeetr}
\bibliography{reference.bib}

\end{document}